\documentclass[mnsc]{informs3}

\OneAndAHalfSpacedXII

\usepackage[sorting=none, maxcitenames=1, maxbibnames=1]{biblatex}
\usepackage{amsmath}
\usepackage{bbm}
\usepackage{booktabs}
\usepackage{multirow}
\usepackage{subcaption}

\TheoremsNumberedThrough     
\ECRepeatTheorems
\EquationsNumberedThrough    

\MANUSCRIPTNO{MS-0001-1922.65}

\begin{document}

\RUNTITLE{}

\TITLE{Competing Risks:\\Impact on Risk Estimation and Algorithmic Fairness}

\ARTICLEAUTHORS{%
\AUTHOR{Vincent Jeanselme}
\AFF{Columbia University, New York\\University of Cambridge, Cambridge\\\EMAIL{vincent.jeanselme@mrc-bsu.cam.ac.uk}}
\AUTHOR{Brian Tom, Jessica Barrett}
\AFF{University of Cambridge, Cambridge}
} 

\ABSTRACT{%
Accurate time-to-event prediction is integral to decision-making, informing medical guidelines, hiring decisions, and resource allocation. Survival analysis — the quantitative framework used to model time-to-event data — accounts for patients who do not experience the event of interest during the study period, known as censored patients. However, many patients experience events that prevent the observation of the outcome of interest. These \emph{competing risks} are often treated as censoring, a practice frequently overlooked due to a limited understanding of its consequences.
Our work theoretically demonstrates why treating competing risks as censoring introduces substantial bias in survival estimates, leading to systematic overestimation of risk and, critically, amplifying disparities. First, we formalize the problem of misclassifying competing risks as censoring and quantify the resulting error in survival estimates. Specifically, we develop a framework to estimate this error and demonstrate the associated implications for predictive performance and algorithmic fairness. Furthermore, we examine how differing risk profiles across demographic groups lead to group-specific errors, potentially exacerbating existing disparities. Our findings, supported by an empirical analysis of cardiovascular management, demonstrate that ignoring competing risks disproportionately impacts the individuals most at risk of these events, potentially accentuating inequity.
By quantifying the error and highlighting the fairness implications of the common practice of considering competing risks as censoring, our work provides a critical insight into the development of survival models: practitioners must account for competing risks to improve accuracy, reduce disparities in risk assessment, and better inform downstream decisions.
}%



\maketitle

\section{Introduction}

Accurate estimation of the time until an event of interest occurs is critical to inform decisions. In marketing, predicting customer churn helps efficient targeting of retention campaigns~\cite{neslin2006defection}. In human resources, forecasting employee turnover informs hiring. In healthcare, estimating the risk of an adverse event, such as a cardiovascular event, supports treatment decisions and the initiation of preventive care~\cite{mangione2022statin}. 

The central challenge in modeling time-to-event outcomes is their partial observation in historical data. During the observation period available for analysis, not all individuals experience the event of interest. For instance, marketers must act using a few months of data after the release of a new product to prevent customer attrition. Similarly, financial costs often limit the length of clinical trials, resulting in patients without the observed outcome of interest. These partial observations are known as \emph{censoring}, where individuals’ outcomes remain unknown beyond the study period.

Survival analysis handles censoring by accounting for the information from individuals who have not yet experienced the event but remain at risk. However, many individuals experience other events that prevent the occurrence of the one of interest. While both censoring and \emph{competing risks} result in an unobserved event, they have fundamentally different implications: censored patients remain at risk beyond the observation period, whereas patients experiencing competing risks will never develop the event of interest. In other words, these individuals have exited the risk set altogether. Table~\ref{tab:examples} illustrates different competing risks that may manifest across domains.
For instance, a client switching to a competitor may still be within the marketer’s strategic purview, while a customer who moves abroad is no longer reachable. Similarly, in healthcare, death from a non-cardiac cause precludes a future cardiac event, and in workforce analytics, retirement or layoff removes individuals from the risk set of voluntary resignation. Our work demonstrates that treating such individuals as censored, a common but flawed practice, leads to biased risk estimates. 

\begin{table}[!ht]
 \centering
      \caption{Decisions informed by survival models and relevant competing risks}
      \small
     \label{tab:examples} 
     \begin{tabular}{ccc}
         \toprule
         \textbf{Decision Context}              & \textbf{Modeled Event}          & \textbf{Competing Risks} \\ \midrule
         Cancer treatment recommendation~\cite{katzman2018deepsurv}         & Cancer recurrence & Death from other cause \\
         Loan issuance~\cite{dirick2017time}         & Default & Repayment, prepayment, death \\
         Customer retention~\cite{braun2011modeling} & Service cancellation & Relocation, death \\
         Hiring~\cite{barrick2009hiring, jin2020rfrsf, zhu2019coxrf} & Resignation & Layoff, disability, retirement \\
         \bottomrule
     \end{tabular}

 \end{table}

Despite their prevalence and practical significance, competing risks are frequently overlooked in applied research~\cite{austin2016introduction, koller2012competing}. In their review of 50 medical research papers investigating time-to-event modeling, \textcite{koller2012competing} identified competing risks among 70\% of the studies, yet more than half inadequately accounted for them in their analysis. The incorrect practice of treating competing risks as censoring~\cite{austin2017accounting} introduces biases in risk estimation, as previously demonstrated empirically across domains~\cite{fisher1974presenting, leung1997censoring, satagopan2004note, schuster2020ignoring, frydman2022random}. However, practitioners continue to overlook this practice as the impact of competing risks on risk estimates has not been theoretically quantified. Our work fills this gap. 

This oversight is not merely a theoretical concern; it has direct repercussions on decision-making. In medicine, misestimated survival risk can result in the overtreatment of low-risk patients and, equivalently, under limited resources, the undertreatment of high-risk individuals. For instance, clinical guidelines for cardiovascular prevention rely on a 10-year risk threshold to recommend statins~\cite{mangione2022statin}. Overestimating this risk due to unmodeled mortality may expose low-risk patients to unnecessary medication and its side effects, while underestimating the risk may delay needed interventions.
Similarly, in customer retention modeling, misestimating churning risk may result in misallocating marketing resources to unreachable individuals. 

While previous research has discussed some of the risks of ignoring competing risks, our work demonstrates that risk misestimation is not uniformly distributed across the population. When groups differ in their incidence of the competing risks, considering these competing risks as censoring introduces group-specific disparities in risk estimates.  In the workforce, women have been historically more likely than men to leave the workforce for caregiving or other non-voluntary reasons~\cite{frederiksen2008gender}, which act as competing risks to voluntary resignation. When survival models fail to account for these group-specific risk profiles, they systematically overestimate risk for groups with higher competing risk prevalence. 

\begin{figure*}[!t]
\centering
    \includegraphics[width=0.8\textwidth]{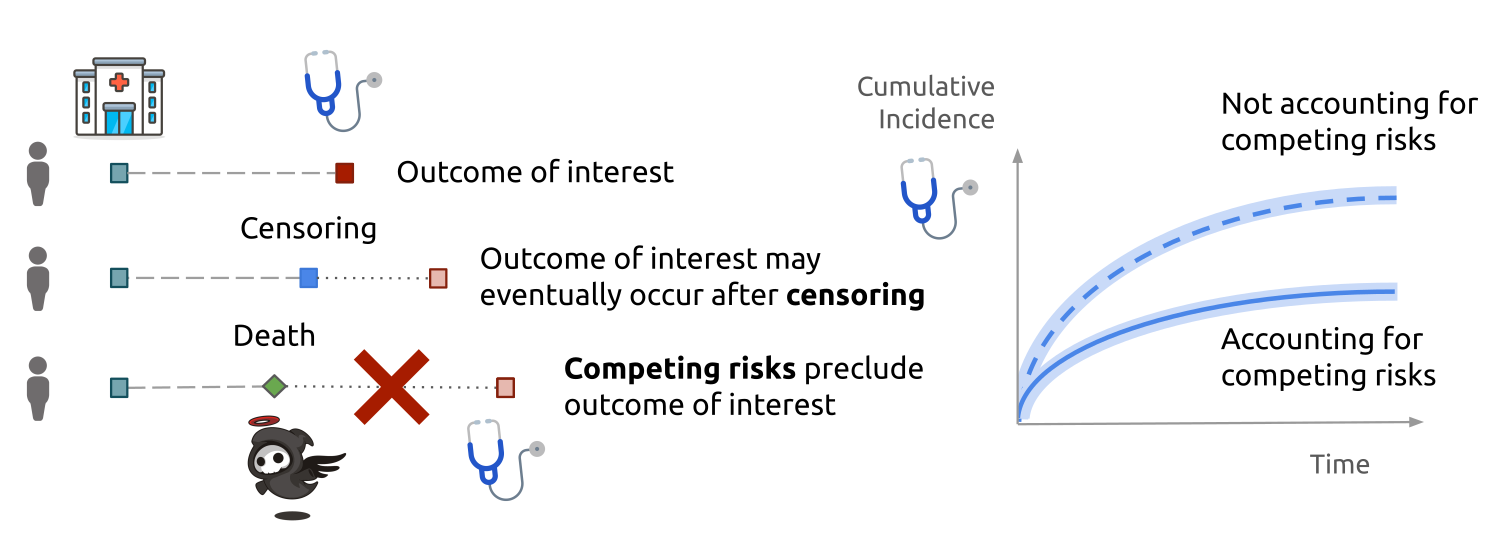}
    \caption{Competing risks preclude the event of interest. Modeling competing risks as censored events biases predictions, impacting downstream decisions.}
    \label{fig:motivation}
\end{figure*}

This paper demonstrates that ignoring competing risks not only biases survival estimates but introduces systematic performance gaps across subgroups, with real-world implications for decision-making. To reach this conclusion, our contributions are three-fold:
\begin{itemize}
    \item We develop a theoretical framework that characterizes the bias in survival estimates caused by misclassifying competing risks as censoring.
    \item We show that this bias manifests unequally across groups with different competing risk profiles, producing systematic error gaps that raise fairness concerns.
    \item We validate these findings through a real-world study on cardiac event prediction.
\end{itemize}

Our contributions have important managerial implications. Risk estimates inform high-stakes decisions --- from medical treatment to workforce management. Failing to account for competing risks in these risk estimation models diminishes their predictive utility and exacerbates inequities between groups with different risk profiles, potentially worsening disparities in downstream decisions, such as access to treatment, promotions, loans, or targeted support. Our theoretical and experimental contributions highlight a simple opportunity for improvement: predictive models must account for competing risks.

The remainder of this paper is structured as follows: Section~\ref{sec:lit} reviews related work on competing risks and algorithmic fairness in survival analysis. Section~\ref{sec:formalisation} formalizes the problem and provides a theoretical analysis of the biases introduced by disregarding competing risks. Section~\ref{sec:simulation} presents simulation results validating these theoretical insights, and Section~\ref{sec:case} studies the empirical impact on a real-world medical dataset.

\section{Related Work}
\label{sec:lit}
While both survival analysis and algorithmic fairness are mature research fields, the problem of competing risks often remains unaccounted for in practice~\cite{wang2019machine, koller2012competing, monterrubio2022review}, and its intersection with algorithmic fairness even less so. This review covers (i) the methodological foundations of survival analysis and competing risks, (ii) the known empirical and theoretical limitations of treating competing risks as censoring, and (iii) the intersection of survival analysis with algorithmic fairness.

\subsection{Survival analysis and competing risks}
Multiple strategies have been proposed to address censoring. Non-parametric estimators such as Kaplan-Meier~\cite{kaplan1958nonparametric} characterize the average survival in the presence of censored patients. Of particular importance in applied research is the Cox model~\cite{cox1972regression}, which assumes the instantaneous risk of observing the event of interest, ie the hazard, to be proportional between patients. This assumption results in a simplified optimization while accounting for covariates influencing the survival outcome. DeepSurv~\cite{katzman2018deepsurv} extends this model using a neural network for increased flexibility in the proportional deviation from the non-parametric mean survival. Further machine learning strategies have been proposed to account for censoring while relaxing these assumptions~\cite{rindt2022survival, jeanselme2022neural, jeanselme2024language, nagpal2021deep, nagpal2020deep}. 

A second challenge that often occurs is the presence of competing risks ---events precluding the one of interest. Fine-Gray~\cite{fine1999proportional} proposes a parallel to the Cox model for the joint risk of observing an event of interest. 
The machine learning literature has introduced methodologies to further relax this assumption while accounting for competing risks through ordinary differential equation~\cite{danks2022derivative}, time discretisation~\cite{lee2018deephit} or the use of monotonic neural networks~\cite{jeanselme2023neural}.

Despite a growing methodological toolkit, applied analyses often default to Cox models or non-parametric estimators such as Kaplan-Meier, ignoring competing risks~\cite{koller2012competing, monterrubio2022review}, as seen in domains like cardiovascular risk prediction~\cite{wilson1998prediction}, or employee attrition modeling~\cite{zhu2019coxrf}.

\subsection{Known shortcomings of not accounting for competing risks}
The literature emphasizes the inadequacy of the non-parametric Kaplan-Meier~\cite{kaplan1958nonparametric} survival estimate under competing risks with multiple empirical evidence~\cite{gooley1999estimation, southern2006kaplan, wolbers2014competing}. \textcite{satagopan2004note} provides intuitions on the biases resulting from considering competing risks as censoring: the estimated risk for individuals experiencing the event of interest is overinflated using Kaplan-Meier estimates because censored individuals remain at risk while experiencing competing risks should reduce this set. Despite this evidence, Kaplan-Meier estimates continue to be misapplied~\cite{van2016competing}, as demonstrated by \textcite{van2016competing}'s analysis of 100 publications, revealing that nearly half relied on the Kaplan-Meier estimator in the presence of competing risks. Critically, a third of these studies overestimate the underlying risk by more than 10\%~\cite{van2016competing}.

Empirical evidence in the literature also highlights biases when comparing the Cox and Fine-Gray models~\cite{coemans2022bias, schuster2020ignoring, wolkewitz2014interpreting}. \textcite{wolkewitz2014interpreting} advocates for careful examination of assumptions underlying survival models, evidencing that excluding competing risks in the Cox model results in overestimating risk. \textcite{ wolbers2009prognostic, elmer2023time, berry2010competing, } echo this caution, emphasizing the importance of considering competing risks. Particularly, \textcite{wolbers2009prognostic, berry2010competing} underline the impact on frail populations who are at a higher risk for competing risks. This comment highlights the potential group-specific misestimation and its algorithmic fairness consequences studied in our work. 

Despite empirical documentation of the error resulting from inadequate modeling of competing risks, formal quantification of this error is still lacking. Our work aims to fill this gap.

\subsection{Algorithmic fairness and survival analysis}
The growing literature on algorithmic fairness highlights the risks of deploying models that encode or amplify inequities embedded in observed data. In healthcare, \textcite{obermeyer2019dissecting} demonstrates the underestimation of Black patients' healthcare needs when using healthcare cost as a proxy to train a model to predict healthcare utilization. In hiring, predictive models trained on historical decisions to predict applicants' fit for a job have been shown to result in underestimating women's qualifications~\cite{dastin2022amazon}. 

Although the literature has studied both survival modeling and algorithmic fairness, there is limited research on the intersection between these domains. Existing works at this intersection focus on group disparities associated with censoring. Specifically, the non-observation of events complicates the quantification and mitigation of group disparities. \textcite{zhang2022longitudinal} introduce a concordance impurity metric to quantify group differences in discriminative performance for censored data. Leveraging this metric, the authors alter survival random forests with an updated splitting criterion to ensure fairer survival modeling. \textcite{hufairness} approach the problem through a distributionally robust optimization, improving algorithmic fairness properties without access to demographics. \textcite{do2023fair} explore a training procedure minimizing the inter-group mutual information. Instead of modifying the model's training, \textcite{zhao2023fairness} propose pre- and post-processing to maximize algorithmic fairness. More recently, \textcite{rahman2022fair} propose a pseudo-value model to reduce the bias resulting from ignoring the censoring mechanisms, particularly when assumptions of non-informativeness do not hold. 

No prior study has analyzed the impact of the common practice of modeling competing risks as censoring on algorithmic fairness.
Our work contributes to this literature by demonstrating that, even in the absence of bias in the data, ignoring observed outcomes may result in differential risk estimates.

\section{Considering competing risks as censoring}
\label{sec:formalisation}
This section formalizes the concept of competing risks and demonstrates how modeling them as censoring biases risk estimates and results in group-specific errors. 

\subsection{Competing risks}
\begin{definition}[Competing risks]
    A competing risk is an event that prevents the occurrence or observation of the event of interest. Contrary to censoring, a competing risk is informative: its occurrence implies that the event of interest can no longer occur.
\end{definition}

Consider $X$, the observed covariates, $G$, the group membership, $T$, the time of the first observed event, and $D$, its associated event type. These are defined by:
$$T := \min(C, T_1, ..., T_R), D := \argmin(C, T_1, ..., T_R)$$
with $T_r$ the random variable associated with the time to the realization of event $r \in [\![1, R]\!]$ and $C$ the (right-)censoring time. In this context $D = 0$ if no event occurs before the individual leaves the study, meaning the individual is censored. If $D = r$, the event $r$ is the first to be observed, with $T_r$ the minimum observed time.

Our objective is to estimate the risk of a specific event $r$ \emph{while accounting for all potential competing risks}. Importantly, $T_r$ does not represent this quantity as it corresponds to the occurrence of $r$, \emph{independently} of the other events, ignoring that another event might preclude the occurrence of $r$. 

Correct modeling requires capturing the joint distribution of the time to the first event $T'$ and its associated type $D'$, {independently of the censoring time}:
$$T' := \min(T_1, ..., T_R), D' := \argmin(T_1, ..., T_R)$$

Note that, in the context of a single risk, $D' = 1$ is the only possible outcome, resulting in $T' = T_1$.

\begin{figure}[!ht]
    \centering
    \includegraphics[width=0.25\linewidth]{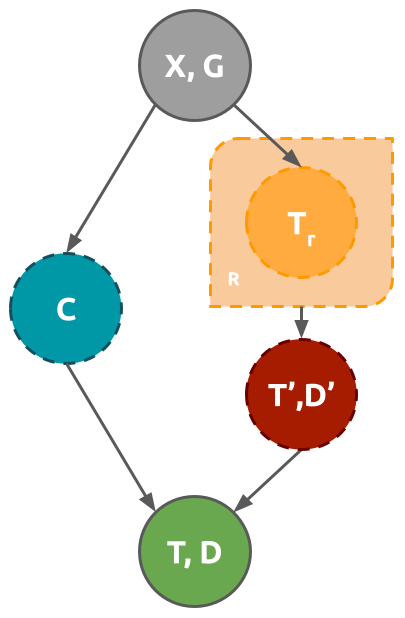}
    \caption{Competing risks {DAG}. \textit{Each node represents a random variable --- a dotted contour indicates that the variable may be unobserved, a solid one, observed. The square around the random variable  $T_r$ denotes the repetition of the variable for $r\in [\![1, R]\!]$. Arrows between nodes represent dependencies.}}
    \label{competing:dag}
\end{figure}

Figure~\ref{competing:dag} illustrates the dependence structure between the previously introduced variables through a directed acyclic graph. In this graph, each arrow describes a dependence between random variables and solid contours correspond to observed variables. $X$, $G$, $D$ and $T$ are observed, while the variables of interest $T'$ and $D'$ are not. The plate notation denotes repeated variables across the $R$ competing risks. Importantly, this representation underlines that $T'$, $D'$ depend on \emph{all} events $T_r$. 

This formalism lets us describe the central quantity modeled when accounting for competing risks. The survival function $S(t)$ is the probability of observing none of the competing risks before time $t$: 
$$S(t) := \mathbb{P}(T' \geq t) = 1 - \sum_{r \in [\![1, R]\!]} F_r(t)$$ where $F_r$ denotes the probability of observing the event $r$ before time $t$ without prior occurrence of any competing event. This quantity, known as the Cumulative Incidence Function (CIF) for the event $r$, can be described as the joint distribution of the previous variables: 
\begin{align}
    \label{eq:cif}
    F_r(t) := \mathbb{P}(T' < t, D' = r)
\end{align}
Our focus is on comparing this correctly specified risk estimate to the common but flawed practice of treating competing risks as censoring.

\subsection{Impact on cumulative incidence estimate}
\label{sec:survival_error}
Treating competing risks as censoring corresponds to ignoring the dependence between events, thereby estimating the marginal distribution of $T_r$ instead of the joint distribution of $T'$ and $D' = r$. Adequate modeling of competing risks consists of evaluating the CIF described in equation~\eqref{eq:cif} --- denoted as $F^C_r$ in this section, whereas ignoring them means estimating the marginal distribution:
$$F^{NC}_r := \mathbb{P}(T_r < t)$$ 

To measure the discrepancy introduced by misspecification, we define the relative cumulative incidence discrepancy. This measure corresponds to the normalized difference in the two estimates, formally:
\begin{definition}[Relative cumulative incidence discrepancy]
The relative cumulative incidence discrepancy associated with the practice of considering competing risks as censoring, denoted by $L^r$:
\begin{align}
    L^r(t, x) := \frac{F^{NC}_r(t\mid x) - F^{C}_r(t\mid x)}{\max(F^{NC}_r(t\mid x), F^C_r(t\mid x))}
    \label{eq:L}
\end{align}
with $r$ the event of interest, $t$, the survival horizon, and $x$, the covariates informing survival.
\end{definition}

\paragraph{Intuition.} This metric quantifies the relative error between the incorrect and correct handling of competing risks. The magnitude of $L^r$ quantifies the disagreement in the probability estimates of experiencing the event $r$ following the two approaches. Positive values denote an overestimation of the risk when considering competing risks as censoring. When null, this implies that both methods yield the same estimate, meaning that accounting for competing risks would not alter the risk estimation. One would not observe predictive gains from modeling competing risks in this case. Negative values indicate an underestimation of the risk when competing risks are treated as censoring.

Using the previous definitions and Bayes theorem, we now derive a simple but critical theorem linking this discrepancy to the probability of observing any event other than $r$.
\begin{theorem}
    \label{thm:estimate_bias}
    The relative cumulative incidence discrepancy for a given outcome $r$ is the probability of observing a competing event prior to $r$ given the covariates $x$:
    \begin{align*}
        L^r(t, x) = \mathbb{P}(D' \neq r \mid T_r < t, x)
    \end{align*}
\end{theorem}

\paragraph{Intuition.} This derivation demonstrates how incorrectly modeling competing risks biases the risk estimate proportionally to how likely an individual is to experience a competing event by time $t$. Note that this expression highlights the positivity of this expression, which substantiates the empirical evidence that treating competing risks as censoring \emph{overestimates} the risk estimate~\cite{fisher1974presenting, leung1997censoring, satagopan2004note, schuster2020ignoring}. 
This occurs because failing to account for competing risks keeps individuals who should have been removed (due to experiencing a competing event) in the risk set, inflating the estimated probability of the event of interest. 
Crucially, this demonstrates that an individual more likely to experience a competing risk derives greater benefit from accounting for competing risks.

\subsection{Impact on group-specific estimate}
\label{sec:gap_survival_error}
While Theorem~\ref{thm:estimate_bias} describes individual-level errors, we now consider group-level effects, which are central to algorithmic fairness. Consider the membership to a protected group $g$ associated with race, sex, or socio-economic factors. Following the \textit{equal performance} definition of fairness introduced in~\cite{rajkomar2018ensuring}, we aim to minimize the difference $\Delta^r_g$ between the previous group-specific discrepancy, defined as follows.

\begin{definition}[Group-specific discrepancy]
The group-specific discrepancy from considering competing risks as censoring is the expected discrepancy $L_r$ across members of a group $g$ for the event of interest $r$ at time $t$:
    $$L^r_g(t) := \mathbb{E}_{x_i \mid g_i = g} \left[L^r(t, x_i)\right]$$
\end{definition}

\begin{definition}[Discrepancy gap]
The gap at a time horizon $t$ from considering competing risks as censoring between members of a group $g$ and the rest of the population is the difference in group-specific discrepancies:
    $$\Delta^r_g(t) := L^r_g(t) - L^r_{\neg g}(t)$$
\end{definition}

\paragraph{Intuition.} The discrepancy gap quantifies how the common practice of treating competing risks as censoring differentially impacts groups. The group-specific discrepancies span the range $[\![-1, 1]\!]$, where $\Delta^r_g(t) = 0$ indicates that the error associated with incorrect competing risk is the same across groups, implying no differential impact. Positive values correspond to larger discrepancies for group $g$, meaning that this group is more affected by the misestimation than the rest of the population. Conversely, negative values indicate that $g$ is less impacted by the misestimation compared to the rest of the population.

These definitions and Theorem~\ref{thm:estimate_bias} directly lead to the following expression of the discrepancy gap.

\begin{lemma}[Fairness gap resulting from ignoring competing risks]
\label{thm:fairness}
The gap from considering competing risks as censoring between members of a group $g$ and the rest of the population is the difference in the risk for any of the competing risks other than $r$:
$$\Delta_g^r(t) = \mathbb{P}(D' \neq r \mid T_r < t, g) -  \mathbb{P}(D' \neq r \mid T_r < t, \neg g)$$
\end{lemma}

\paragraph{Intuition.} The fairness gap arises directly from group-level differences in the likelihood of experiencing competing risks. If individuals in group $g$ are more likely to face competing events than the complementary group, failing to account for these risks introduces systematic bias in survival estimates, disproportionately affecting the group with the higher competing risk rate. 

Crucially, considering more flexible strategies would not resolve this issue. No matter how expressive a survival model is, it propagates differential estimates' errors across groups if it does not correctly account for competing risks. Therefore, ensuring fairness in survival predictions and downstream decisions requires explicitly modeling competing risks rather than considering more flexible predictive methods.

\section{Empirical evidence of the impact of different competing risks handling strategies}
\label{sec:simulation}
To empirically validate the theoretical biases introduced by improperly accounting for competing risks, we design a simulation study where the underlying survival and each event distribution are known. This setting allows us to quantify the estimation errors induced by considering competing risks as censoring and to assess whether more flexible modeling strategies would benefit differently from accounting for competing risks.

We first describe the data generation process and then evaluate several survival modeling strategies --- both classical and state-of-the-art neural networks --- by comparing their empirical errors to theoretical expectations across 25 replications of the previous simulation.

\subsection{Data generation}
\label{simulation}
Our simulations rely on a synthetic population of $N = 30,000$ individuals with $10$ associated covariates $X \in \mathbb{R}^{10}$, group membership $G \in \{0, 1\}$ and associated time and cause of event $T, D$. This setting is not meant to mimic any specific real-world application but instead allows us to precisely study how modeling choices affect the alignment between estimated and true CIFs. 

The data generation follows a three-step process, adapted from prior works~\cite{lee2018deephit, nagpal2022counterfactual} summarized below and detailed in Appendix~\ref{app:data_generation}.
\paragraph{Covariates and group membership.} Group membership $G$ is drawn from a Bernoulli distribution with equal probability. Covariates $X$ are then sampled from normal distributions with unit variance and group-specific means, inducing group-specific distributions in the covariates.

\paragraph{Events generation.} Each individual is at risk of two competing events. The cause-specific hazards follow Gompertz distributions with a shared scale parameter but distinct shape parameters dependent on individual covariates, ensuring closed-form expressions for all CIFs. 

\paragraph{Censoring.} Censoring times are generated independently of the covariates and event mechanisms, also following a Gompertz distribution. Observed event times and types ($T, D$) correspond to the minimum between the first observed competing risk time and censoring.

Using this procedure, we generate 25 datasets. For each, our objective is to model the survival distribution associated with the competing risk 1, relying on the observed $T, D, X, G$. While the number of simulations may appear limited due to the prohibitive computational cost of training each of these methods, this number improves performance quantification compared to the common reliance on a unique synthetic dataset in the machine learning literature. To improve reproducibility, the code to generate the data and compare the different models is made available on GitHub\footnote{\url{https://github.com/Jeanselme/CompetingRiskFairness}}.

\subsection{Empirical settings}
\label{sec:exp}
We compare a range of survival models, from classical statistical methods to recent deep learning approaches, to assess whether and how their treatment of competing risks affects empirical performance and alignment with theory.

\paragraph{Models.} First, we considered the well-established \textbf{Fine-Gray} model~\cite{fine1999proportional}. This model extends the traditional Cox model~\cite{cox1972regression} to competing risks. Specifically, it models the sub-distribution hazards $h_r$, a quantity directly linked to CIF as defined below, in a similar way as the Cox model models the hazard.
$$h_r(t\mid x) := - \frac{\partial \log(1 - F_r(u \mid x))}{\partial u}\bigg|_{u = t}$$
Under a similar proportionality assumption as the Cox model, the Fine-Gray assumes the sub-distribution hazard to have the form:
$$h_r(t\mid x) = h_{r, 0}(t) e^{\beta x}$$
with $\beta$, the parameter to estimate and $h_{r, 0}(t)$ a baseline sub-distribution hazard, usually estimated through a non-parametric estimator. 

Additionally, we compare with state-of-the-art neural network approaches that account for competing risks: Neural Fine-Gray (\textbf{NeuralFG},~\cite{jeanselme2023neural}), \textbf{DeepHit}~\cite{lee2018deephit} and \textbf{DeSurv}~\cite{danks2022derivative}.
These strategies estimate the CIF while avoiding the proportionality assumption made by the Fine-Gray model. DeepHit discretizes the time horizon for modeling the survival outcome as a discrete classification problem. DeSurv models the CIF by solving a differential equation through numerical integration. NeuralFG proposes to use monotonic neural networks to directly satisfy the properties of the CIF and avoid the previous approximations. These different strategies, therefore, present improved flexibility in comparison to the traditional modeling strategies. Further, avoiding approximations as proposed in NeuralFG has been shown to improve predictive capacity~\cite{jeanselme2023neural}. 

Finally, as our aim is to measure the gain of these methodologies over the common handling of competing risks as censoring, we considered each model's non-competing alternative (No CR.) that treats competing risks as censoring. Practically, this consists of replacing any observed event other than the one of interest with a censoring indicator. Note that the non-competing alternative to Fine-Gray\cite{fine1999proportional} simplifies as the common Cox model~\cite{cox1972regression}.

\paragraph{Training procedure.} We analyze the performance of these different strategies over the 25 repetitions of the previously described simulations. For each simulated dataset, we split the data into two: 80\% for development and the rest for testing. Following standard machine learning practices, we further divide the development set into 10\% for hyperparameter tuning, 10\% for early stopping, and the rest for training. For all neural network strategies, we use random search on the following hyperparameter grid over 100 random iterations: learning rate ($10^{-3}$ or $10^{-4}$), batch size ($100$, $250$), dropout rate ($0$, $0.25$, $0.5$ or $0.75$), number of layers ($[\![1, 4]\!]$) and nodes ($25$ or $50$). All models are optimized using an Adam optimizer~\cite{kingma2014adam} over $1,000$ epochs with early stopping if the loss increases for three consecutive epochs. Additionally, for DeSurv, we followed the original paper's recommendation of a 15-point Gauss-Legendre quadrature to integrate the {CIF}s. Similarly, DeepHit uses a regular discretization of the time horizon in $15$ bins.

\subsection{Alignment between theory and empirical discrepancy}
\label{verify_theory}
The simulations enable a direct comparison between theoretical and empirical results. For each individual and model, we compute the absolute difference between the estimated CIF and the true, known, CIF for the outcome of interest. We then measure the difference observed between the methodologies considering competing risks and their non-competing alternative over the 25 simulations. Specifically, we evaluate the empirical relative discrepancies, defined as:
$$\hat{L^1}(t, x) := \frac{\hat{F}^{NC}_1(t\mid x) - \hat{F}^{C}_1(t\mid x)}{\max(\hat{F}^{NC}_1(t\mid x), \hat{F}^{C}_1(t\mid x))}$$

\begin{figure}[!ht]
    \centering
    \includegraphics[width=0.9\textwidth]{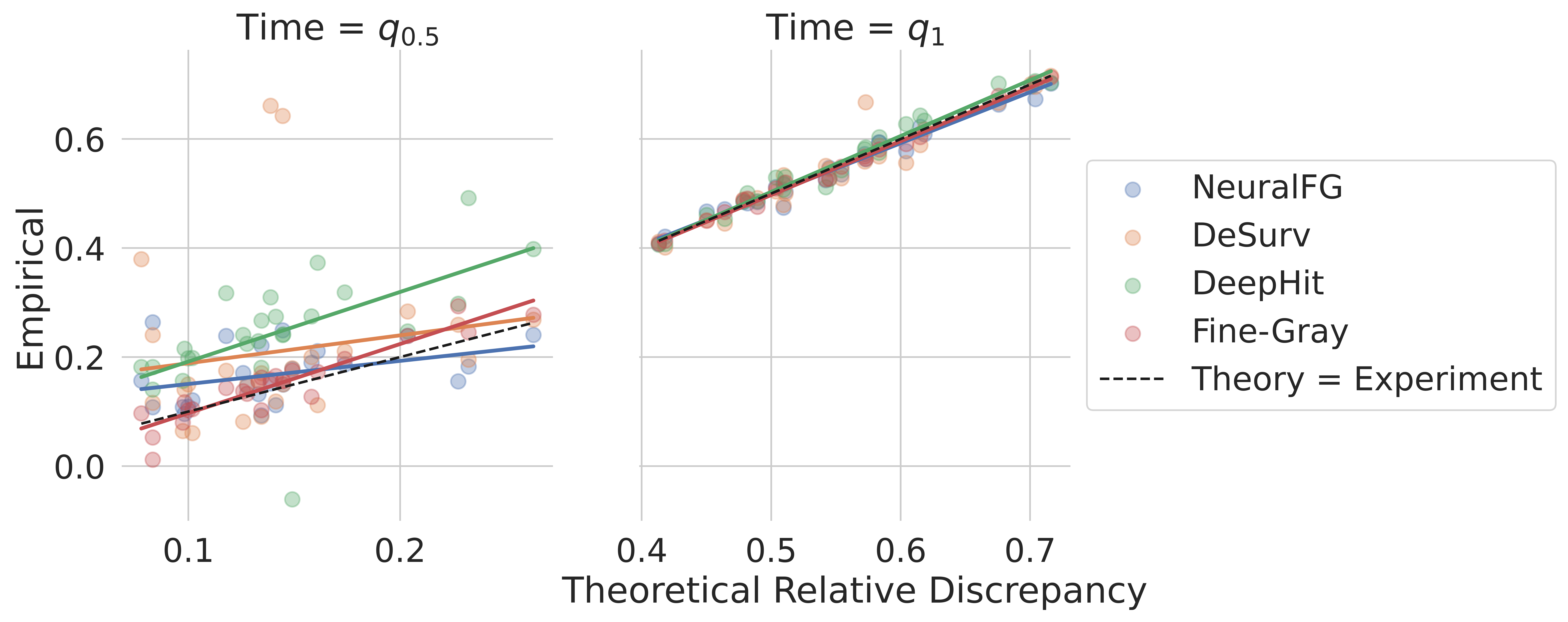}
    \caption{Theoretical and empirical relative discrepancy (${L^1}$) computed for each simulation and method at evaluation times $q_{0.5}$ and $q_{1}$. Each line represents the linear regression fit associated with a method's discrepancy across simulations. The closer to the dashed theoretical line, the more aligned the methodology is with the theory. The larger the value is, the larger the gain of considering competing risks is.}
    \label{fig:relative}
\end{figure}

Figure~\ref{fig:relative} displays this observed quantity on the y-axis as a function of the theoretical one on the x-axis at the 0.5 and 1 quantiles of the observed event times ($q_{0.5}$ and $q_{1}$). In this controlled setting, the average theoretical relative discrepancy is estimated as $\mathbb{E}_x [T_2 < T_1 \mid T_1 < t, x]$ (see Appendix~\ref{apd:derivation} for derivation). In this figure, each point is a simulation; each line is a regression across simulations for a method. Closer alignment to the diagonal reflects enhanced agreement with theoretical estimates. Table~\ref{table:ccc} further presents the root mean squared error (RMSE) between empirical error and theory, numerically quantifying the error between these quantities for each methodology.

\begin{table}[!ht]
\small
    \centering
    \setlength\extrarowheight{-5pt}
    \begin{tabular}{rcccc}
        \toprule
           & \multicolumn{2}{c}{RMSE$(L_1)$} & \multicolumn{2}{c}{RMSE$(\Delta_g)$} \\
        \textbf{Model} &         $q_{0.5}$ &    $q_{1}$&         $q_{0.5}$ &    $q_{1}$\\
        \cmidrule(lr){2-3}\cmidrule(lr){4-5}
        NeuralFG & 0.059 (0.012) & 0.014 (0.002) & 0.095 (0.015) & 0.027 (0.003) \\
        DeSurv & 0.159 (0.047) & 0.024 (0.006) & \textbf{0.091} (0.024) & \textbf{0.022} (0.003) \\
        DeepHit & 0.130 (0.012) & 0.016 (0.001) & 0.167 (0.020) & 0.119 (0.012) \\
        Fine-Gray & \textbf{0.027} (0.004) & \textbf{0.009} (0.001) & 0.191 (0.023) & 0.076 (0.011) \\
        \bottomrule
    \end{tabular}
    \caption{Root mean squared error (bootstrapped standard deviation) between empirical and theoretical quantities measuring the gain associated with modeling competing risks.\\
    The smallest error is highlighted in \textbf{bold}.}
    \label{table:ccc}
\end{table}

As illustrated in Figure~\ref{fig:relative} and shown in Table~\ref{table:ccc}, all methods show strong alignment with theory, confirming that inadequate modeling of competing risks results in a risk overestimation proportional to the incidence of the competing event. Critically, increased flexibility, as provided by neural network approaches, does not mitigate this bias. 

Note that the alignment with theory is noisier at earlier times, at which the probability of experiencing an event is near null, resulting in unstable discrepancy estimation. While expected, this caveat highlights that the benefit of proper handling of competing risks may be higher at longer time horizons.

\subsection{Group-level discrepancy}
\begin{figure}[!ht]
    \centering
    \includegraphics[width=0.9\textwidth]{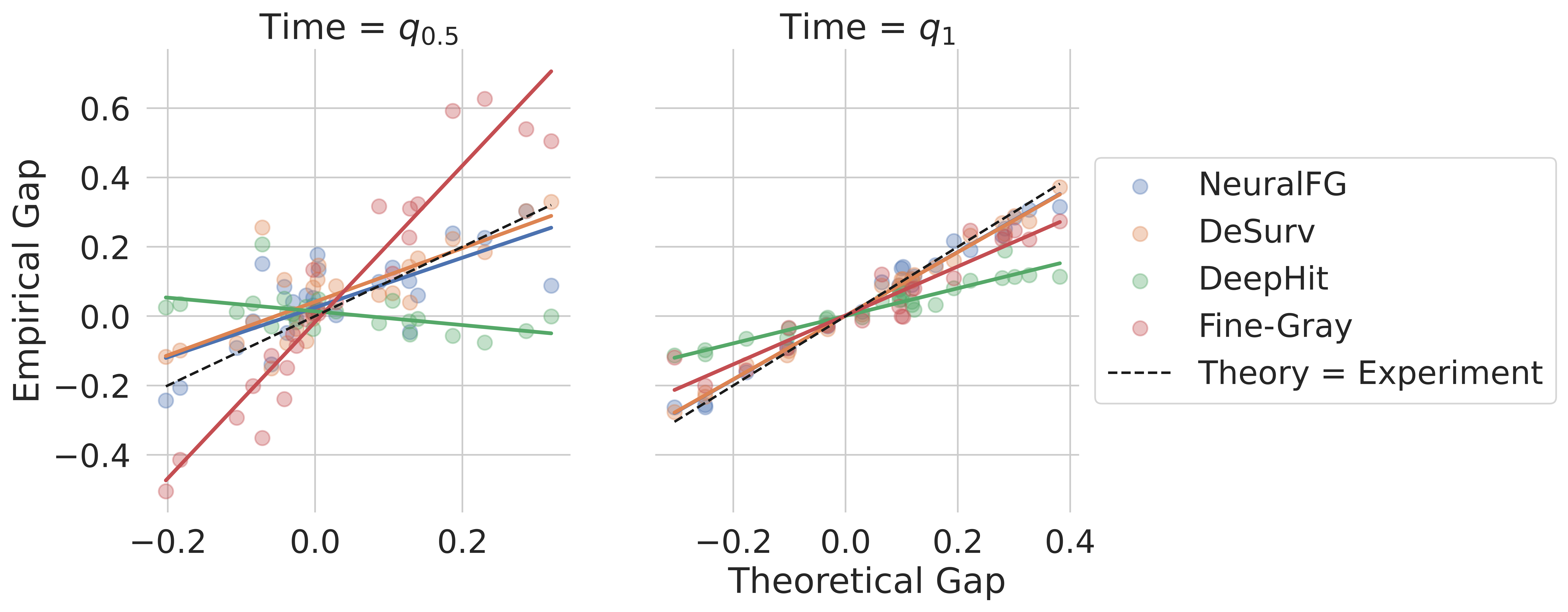}
    \caption{Comparison of the theoretical and empirical gap in relative cumulative incidence discrepancy ($\Delta_g$) computed for each simulation and method at evaluation times $q_{0.5}$ and $q_{1}$. The closer to the dashed theoretical line, the more aligned the methodology is with the theory. More extreme values reflect a larger algorithmic fairness gap resulting from considering competing risks as censoring.}
    \label{fig:gap}
\end{figure}

Knowing group membership and survival distributions in these simulations allows us to verify the theoretical inter-group discrepancy gap introduced in Lemma~\ref{thm:fairness}. Figure~\ref{fig:gap} presents the empirical and theoretical gaps in performance between groups defined by $G$ at the two time quantiles. The x-axis displays the theoretical gap between the group $g$ and the rest of the population computed as:
$$\Delta_g(t) = \mathbb{E}_{x \mid g}[L^1(t, x)] - \mathbb{E}_{x \mid \neg g}[L^1(t, x)]$$

This quantity reflects how groups with different propensities of experiencing the competing risk are impacted by improper competing risk modeling. Theoretically, the more extreme the theoretical gap is, the more pronounced the algorithmic fairness gap benefits from the proper handling of competing risks.

The comparison of the predicted and empirical fairness gap shows the direct impact of competing risks on group-specific error. In Figure~\ref{fig:gap}, empirical and theoretical fairness gaps align, validating the theoretical finding that the larger the difference in experiencing competing risks, the larger the gain from modeling competing risks. This confirms that treating competing risks as censoring can amplify group disparities.

Fine-Gray modeling leads to larger fairness gaps than theoretically estimated, as its underlying proportionality assumption may limit its ability to adapt to group-specific differences. The more flexible neural network models better match group-specific patterns and reduce this discrepancy with theory. Importantly, this observation shows that the fairness gap is \emph{more predictable but not erased} when considering flexible modeling strategies. Only the modeling of competing risks can correct for the associated fairness gap.


\section{Case study: The impact of competing risks on cardiovascular risk management.}
\label{sec:case}
We have theoretically demonstrated the importance of proper modeling of competing risks for improved survival modeling and reduced group-specific errors. In this section, we illustrate the practical implications of these findings in a high-stakes clinical context. Specifically, we study the \textsc{Framingham} Heart Study dataset, a foundational dataset in cardiovascular epidemiology that is central to current cardiovascular management. We explore how survival models mishandling competing risks may impact treatment recommendations. 

\subsection{Empirical setting}

\paragraph{Dataset.} The epidemiological \textsc{Framingham} Heart Study~\cite{kannel1979diabetes} aimed to quantify the impact of risk factors on the propensity of observing cardiovascular disease ({CVD}). Patients were followed for over 20 years, and recurrent questionnaires and clinical measurements such as blood tests, lung function, and treatment were regularly collected. Our analysis focuses on the subset of covariates obtained at the first medical appointment for 4,434 patients to model {CVD} risk. As multiple patients did not observe the primary event of interest and died from other causes over the study period, we consider this outcome as the competing risk of interest.

\begin{table}[!h]
    \small
    \centering
    \caption{Observed outcomes by the different evaluation horizons.}
    \label{tab:quantiles}
    \begin{tabular}{rccc}
            \toprule
           &  \multicolumn{3}{c}{\textbf{Time Horizon}}  \\
           &  1 year &           5 years &           10 years  \\\cmidrule(lr){2-4}
          Censoring & 0.00 \% & 0.00\% & 0.00\% \\
          Death & 6.48\% & 22.27\% & 30.63\%  \\
          CVD & 93.52\% & 77.73\% & 69.37\%\\
          \bottomrule
    \end{tabular}
\end{table}

\paragraph{Evaluation.} To train and evaluate the models, we use a 5-fold cross-validation using the same training procedure as in Section~\ref{sec:exp}. As in this real-world setting, we can not measure the discrepancies to the underlying survival distributions, we instead measure performance employing calibration and discrimination metrics computed at the dataset-specific quantiles of the uncensored population event times. Specifically, we rely on the time-dependent Brier score (td-Brier) and C-index (td-CI) with correction for accounting for competing risks as described in~\cite{van2022validation}. td-Brier measures the alignment between the predicted and observed risks for an event, while td-CI measures the ranking between patients' risks given their observed event time (see \cite{van2022validation} for formal definition). These metrics, therefore, are proxies to the recovery of the CIF. Table~\ref{tab:quantiles} presents the observed outcomes corresponding to three horizons used for evaluation. Note that no censoring is observed before the end of the study, resulting in no censoring at the three horizons of interest. Additionally, a difference with simulation is the presence of missing data. As the impact of missingness on performance is out of the scope of this analysis, we mean imputed all missing values.

\subsection{Predictive performance}
In this section, we first evaluate how competing risk modeling affects the overall predictive performances by comparing the different strategies introduced in Section~\ref{sec:exp}. Our results echo our simulation, demonstrating the predictive gain associated with competing risk modeling. Table~\ref{nfg:tab:result} summarizes the calibration and discriminative performance of the analyzed models on CVD averaged across the 5-fold cross-validation. 

\begin{table*}[!ht]
\small
        \centering
        \begin{tabular}{c c ccc ccc}
            \toprule
              &\multirow{2}{*}{\textbf{Model}} & \multicolumn{3}{c}{\textbf{td-CI} \textit{(Larger is better)}} & \multicolumn{3}{c}{\textbf{td-Brier} \textit{(Smaller is better)}} \\
              &  &           1 year&           5 years&           10 years&           1 year&           5 years&           10 years  \\\cmidrule(lr){3-5}\cmidrule(lr){6-8}
             \parbox[t]{2mm}{\multirow{4}{*}{\rotatebox[origin=c]{90}{Competing}}}& NeuralFG & \textbf{0.958} (0.016) & \textbf{0.882} (0.029) & \textbf{0.830} (0.031) & \textbf{0.019} (0.004) & \textbf{0.042} (0.003) & \textbf{0.077} (0.007) \\
              & DeSurv & 0.940 (0.024) & 0.834 (0.032) & 0.765 (0.030) & 0.029 (0.004) & 0.046 (0.004) & 0.098 (0.006) \\
              & DeepHit & 0.943 (0.026) & 0.860 (0.039) & 0.801 (0.029) & 0.020 (0.003) & 0.045 (0.002) & 0.081 (0.005) \\
              & Fine-Gray & 0.906 (0.026) & 0.847 (0.023) & 0.806 (0.028) & 0.034 (0.007) & 0.054 (0.004) & 0.085 (0.005) \\\cmidrule(lr){3-5}\cmidrule(lr){6-8}
              \parbox[t]{2mm}{\multirow{4}{*}{\rotatebox[origin=c]{90}{No-CR.}}}& NeuralFG &0.949 (0.021) & 0.878 (0.032) & 0.829 (0.035) & 0.032 (0.007) & 0.044 (0.003) & \textbf{0.077} (0.008) \\
              & DeSurv & 0.942 (0.018) & 0.854 (0.020) & 0.789 (0.024) & 0.026 (0.005) & 0.059 (0.008) & 0.138 (0.019) \\
              & DeepHit & 0.946 (0.019) & 0.857 (0.034) & 0.789 (0.039) & 0.023 (0.005) & 0.048 (0.002) & 0.085 (0.007) \\
              & Cox & 0.909 (0.023) & 0.848 (0.022) & 0.809 (0.026) & 0.033 (0.007) & 0.053 (0.004) & 0.084 (0.005) \\
              \bottomrule
        \end{tabular}
    \caption{Comparison of model performance by means (standard deviations) across 5-fold cross-validation on the primary outcome. Best performances are in \textbf{bold}.}
    \label{nfg:tab:result}
\end{table*}

Echoing previous empirical results~\cite{fisher1974presenting, leung1997censoring, satagopan2004note, schuster2020ignoring}, proper modeling of competing risks improves performance compared to the models' non-competing alternative (No-CR). Particularly, the model calibration consistently improves when comparing the two strategies. The only exception to this result is the Fine-Gray model compared to Cox. These models' linearity and proportional hazard assumptions limit their modeling flexibility, resulting in no apparent gain from competing risk modeling and even lower predictive performance.
Finally, the more flexible neural networks result in consistent calibration gain from competing risk modeling but no gain from a discriminative point of view. This observation means that the probability estimate is better aligned when considering competing risks, while the ranking between patients presents limited differences between the different modeling strategies. Critically, the strategy with the least approximation of the likelihood, NeuralFG, results in the best predictive performances. 

\subsection{Impact on group-specific performance}
\label{subsec:disparity}
Beyond average performance, we now examine how mishandling competing risks affects subpopulations differently. As demonstrated in Lemma~\ref{thm:fairness}, patients who are the most likely to experience competing risks benefit the most from this modeling. As cardiovascular risk profiles change with age and sex, we propose to explore the impact of modeling competing risks on these groups. Table~\ref{tab:group} describes the distributions of the outcomes of interest for each group of interest by the end of the observation period. Note that the 50-year-old threshold was selected to divide the population in two. For each group, we then compute the difference in calibration when accounting for competing risks or treating them as censoring using the most accurate predictive model in our analysis: NeuralFG. Additionally, we compute the change in fairness gap defined in Lemma~\ref{thm:fairness}, reflective of how competing risk modeling would impact the predictive gap between groups. Table~\ref{tab:age} (resp. Table~\ref{tab:sex}) summarizes the calibration differences for the different age groups (resp. sex). We defer the analysis of discrimination performance differences to Appendix~\ref{apd:discrimnation_diff}, as they reflect similar conclusions.

\begin{table}[!h]
    \small
    \centering
    \caption{Observed outcomes stratified by age and sex.}
    \label{tab:group}
    \begin{tabular}{rcccc}
            \toprule
           &  \multicolumn{4}{c}{\textbf{Population split}}  \\
           &  $<$ 50 ($n = 2,393$)&           $\geq$ 50 ($n = 2,041$)&           Men ($n = 1,944$) & Women ($n = 2,490$)\\\cmidrule(lr){2-3}\cmidrule(lr){4-5}
          Censoring & 71.29\% & 38.41\% & 45.78\% & 64.26\%\\
          Death & 11.03\% & 25.62\% & 18.93\% & 16.83\%\\
          CVD & 17.68\% &  35.96\% &  35.29\% & 18.92\%\\
          \bottomrule
    \end{tabular}
\end{table}

Older patients have a higher likelihood of death than younger ones. In our datasets, patients above 50 have more than twice the risk of death (25.6\% against 11.0\%), and twice the risk for CVD (40.0\% against 17.7\%). These groups, therefore, present different propensities for the different outcomes, aligning with the theoretical setting most likely to benefit from competing risk modeling.
Table~\ref{tab:age} illustrates this benefit. Both age groups benefit from competing risk modeling, as shown by negative values at each time horizon. However, older patients benefit the most from modeling death as a competing risk, as shown by larger absolute differences. Importantly, we compute the change in algorithmic fairness gap  $|\Delta_{\geq}^C| - |\Delta_{\geq}^{NC}|$ in the last row. If negative, this quantity reflects a reduction in the fairness gap between younger and older patients when accounting for competing risks. Our results shows a consistent reduction in the performance gap between groups when modeling competing risks.

\begin{table}[!ht]
        \small
        \centering
        \caption{Calibration differences - Means and standard deviations over 5-fold cross-validation. {Smaller values correspond to improved calibration of the competing risk model over the non-competing one.} $|\Delta_{\geq}^C| - |\Delta_{\geq}^{NC}|$ corresponds to the difference in the fairness gap between younger and older than 50 patients when modeling competing risks.}
        \label{tab:age}
        \begin{tabular}{cccc}
            \toprule
            \multirow{2}{*}{\textbf{Age}}  & \multicolumn{3}{c}{\textbf{td-Brier Difference}}\\
               &            1 year&           5 years&           10 years \\\cmidrule(lr){2-4}
                $<$ 50 & -0.005 (0.006) & -0.001 (0.002) & -0.000 (0.001) \\
                $\geq$ 50  & -0.022 (0.011) & -0.004 (0.004) & -0.001 (0.003) \\
                \cmidrule(lr){2-4}
                $|\Delta_{\geq}^C| - |\Delta_{\geq}^{NC}|$ & -0.017 (0.009) & -0.002 (0.004) & -0.001 (0.002) \\
                \bottomrule
        \end{tabular}
\end{table}

Similarly, we study sex differences. The average age between men and women is similar (50.0 for women and 49.8 for men). However, women in the United States generally have longer life expectancy~\cite{seifarth2012sex}, rendering them less likely to die over a fixed term study in comparison to men of the same age. We observe this phenomenon in the dataset with a sex-specific risk of death from other causes (16.8\% for women, against 18.9\% for men). While non-significant, this difference in the competing risk mechanism could result in differential gains when modeling competing risks. Table~\ref{tab:sex} confirms this intuition with the difference in Brier score between the competing and non-competing models doubling for men. The fairness gap $|\Delta_{M}^C| - |\Delta_{M}^{NC}|$ aligns with our theoretical results, confirming that modeling competing risks reduces the algorithmic fairness gap when the competing risk's mechanism is group-dependent. 

\begin{table}[!ht]
        \small
        \centering
        \caption{Calibration differences between sex - Means and standard deviations over 5-fold cross-validation. Smaller values correspond to improved calibration of the competing risk model over the non-competing one. $|\Delta_{M}^C| - |\Delta_{M}^{NC}|$ corresponds to the difference in the fairness gap between men and women when modeling competing risks.}
        \label{tab:sex}
        \begin{tabular}{c ccc}
            \toprule
            \multirow{2}{*}{\textbf{Sex}}  & \multicolumn{3}{c}{\textbf{td-Brier Difference}}\\
               &            1 year&           5 years&           10 years \\\cmidrule(lr){2-4}
                Male & -0.020 (0.012) & -0.004 (0.003) & -0.001 (0.002) \\
                Female & -0.007 (0.004) & -0.001 (0.002) & -0.001 (0.003)  \\\cmidrule(lr){2-4}
                $|\Delta_{M}^C| - |\Delta_{M}^{NC}|$ & -0.013 (0.008) & -0.003 (0.003) & 0.000 (0.003) \\
            \bottomrule
        \end{tabular}
\end{table}

\subsection{Impact on practice}
The \textsc{Framingham} dataset was used to derive the eponymous 10-year cardiovascular disease ({CVD}) risk score~\cite{wilson1998prediction}. This score guides clinical practice in preventatively treating patients, usually with a combination of cholesterol-lowering therapy,~e.g.,~statins, and holistic treatment of other {CVD} risk factors~\cite{bosomworth2011practical, d2008general}. To minimize overtreatment and adverse side effects, accurate risk estimates are critical for targeting the population most at risk so as to maximize the benefit-risk ratio~\cite{mangione2022statin}. However, cardiovascular risk scores often rely on a Cox model, which does not account for competing risks~\cite{mangione2022statin, d2008general, van2014performance}.

\begin{figure}
    \centering
    \includegraphics[width=\linewidth]{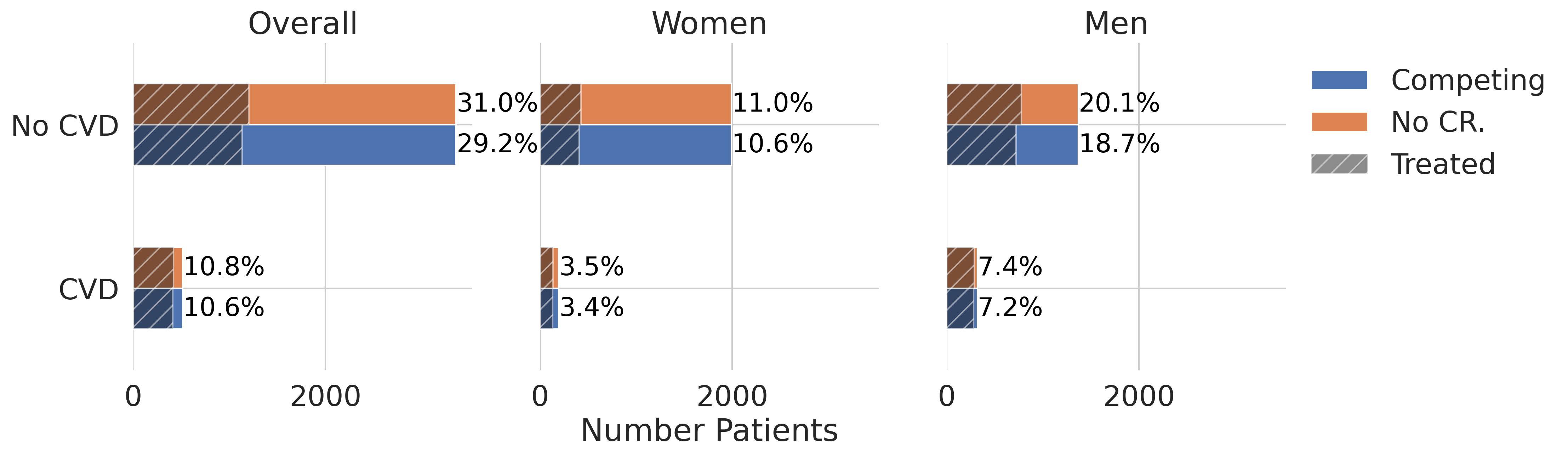}
    \caption{Percentage of treated patients under competing and non-competing risk scores stratified by outcomes for patients older than 40 (overall and stratified by sex).}
    \label{fig:risk}
\end{figure}

Clinical treatment often relies on a discretization of the risk of observing a {CVD} event in the following 10 years~\cite{bosomworth2011practical}. Recent guidelines in the United States suggest placing all patients older than 40 with $\geq 10\%$ risk on cholesterol-lowering drugs~\cite{mangione2022statin}. Furthermore, in the United States alone, several million patients are on these medications~\cite{wall2018vital}. Therefore, even modest shifts in patient risk classification could, at scale, amount to considerable numbers either inappropriately receiving preventative treatment or inappropriately receiving none. To demonstrate how considering competing risks can fundamentally alter such risk profiling, we present in Figure~\ref{fig:risk} the treatment recommendation following~\textcite{mangione2022statin}'s guidelines under both NeuralFG and its non-competing alternative. 

Relying on the competing risk model could reduce the risk of overtreatment in the population. While the percentage of patients not receiving treatment despite being at risk increases by 0.2\% when using the competing risk model, 1.8\% of the population avoids unnecessary treatment.

Furthermore, we stratify the population at risk by sex to examine potential group-specific patterns that may arise from treating competing risks as censoring. As observed in Figure~\ref{fig:risk}, this difference in the competing risk mechanism results in differential gains when modeling competing risks. Specifically, the use of the competing risk models leads to a similar rate of undertreatment (around 0.2\%) for patients with CVD. However, the difference between models results in a risk of overtreatment 3.5 times larger when relying on the non-competing model for men than women. In a resource-constrained environment, such error has a significant impact on treatment allocation, with the risk of treating men more often than women for the same risk. 

In summary, using a non-competing risk score would have important clinical consequences of over- and under-treatment~\cite{schuster2020ignoring}. More predictive models accounting for competing risks must be preferred to ensure better care. While some updated risk assessment tools in Europe, such as SCORE2~\cite{score22021score2}, account for competing risks, this practice remains uncommon. Our study not only echoes the importance of modeling competing risks for improved predictive performance but also underscores its broader fairness implications.

\section{Discussion}

Despite long-standing recommendations in the statistical literature to model competing risks appropriately~\cite{berry2010competing}, many applied fields continue to treat them as censoring~\cite{koller2012competing, monterrubio2022review}. This practice persists partly due to a lack of theoretical understanding and practical tools to quantify its consequences. Our work addresses this gap by formally analyzing the error introduced by ignoring competing risks and empirically validating its impact in real-world settings. Critically, our work highlights the algorithmic fairness implications of this practice, drawing attention to this issue beyond predictive performance.

\subsection{Contributions}

We introduce a framework to characterize the error associated with treating competing risks as censoring. We derive the systematic risk overestimation and demonstrate that this bias is not uniformly distributed: individuals with a higher likelihood of experiencing competing risks systematically suffer from larger errors. These group-dependent errors may cascade through downstream decisions in ways that reinforce or exacerbate existing disparities.

Our empirical findings confirm this concern. In both simulations and a real-world cardiac risk prediction task using the \textsc{Framingham} dataset, we observe consistent overestimation of risk when competing events are misclassified as censoring. Such misestimation may impact downstream decisions, leading to overtreatment, inefficient resource allocation, or the deprioritization of higher-risk individuals in resource-constrained settings.

Our findings are not confined to healthcare. In human resources, conflating resignation with layoffs or retirement results in misestimating retention probabilities and may influence promotion or hiring decisions. In finance, considering loan repayment as censoring leads to misjudged default probabilities and results in unfair lending recommendations. In marketing, ignoring relocation leads to wasted resources and ineffective targeting. Across these domains, failing to model competing risks has a direct impact on predictive performance and fairness, with consequential repercussions in decision-making.

\subsection{Managerial Implications}
Our results have direct implications for decisions based on risk prediction, such as resource allocation or treatment recommendation. Practitioners must not discard other observed outcomes that may inform the one of interest. Across domains where survival models inform high-stakes decisions, failing to collect and model these competing risks may lead to a systematic overestimation of risk. This overestimation can cause harm: inappropriately targeting interventions, misallocating limited resources, or making unjust recommendations.

Our work, therefore, encourages the collection of fine-grained outcome data on why individuals do not experience the outcome of interest, and invites policymakers and model developers to rely on models that adequately account for competing risks. Doing so will improve both the accuracy and fairness of risk predictions, with the potential to reduce disparities in the allocation of support, care, and opportunity.

\subsection{Future directions.} 
While our work quantifies the risks associated with the common practice of treating competing risks as censoring and invites practitioners to account for competing risks, several open questions remain.

To facilitate adoption, future research should explore post-training adjustments of existing risk scores rather than replacing them entirely. Our theoretical results suggest post hoc correction strategies, such as scaling risk predictions by the probability of not experiencing a competing event, as a promising mitigation route for existing tools. These corrections could offer a practical path forward in settings where model replacement is impractical.

Despite the evidenced benefits of competing risk modeling, interpretability remains a challenge. Competing risk models complicate causal interpretations of covariates, often leading practitioners to favor simpler --- but, as demonstrated, biased --- models. Future research should investigate interpretable modeling and post-hoc auditing to improve this dimension.

Finally, while our work treats competing risks as a modeling challenge to improve the estimation of the event of interest, these risks can themselves be informative for downstream decision-making. Different competing risks may call for distinct resolution strategies. For example, in the cardiovascular case study, patients at high risk of death from non-cardiovascular causes might benefit more from palliative interventions than statin therapy. Our future work will explore how modeling competing risks may inform the design of more granular decision processes.

To conclude, our work offers a practical call to action: treating competing risks as censoring is more than a theoretical concern, it is a source of systematic, group-dependent error with potential downstream decision-making repercussions. Proper modeling of competing risks is essential for building predictive systems that are not only more accurate but also more equitable.

\ACKNOWLEDGMENT{The authors acknowledge the partial support of the UKRI Medical Research Council (programme numbers MC\_UU\_00002/5 and MC\_UU\_00002/2 and theme number MC\_UU\_00040/02 – Precision Medicine).}

\printbibliography

\newpage

\begin{APPENDICES}
\section{Proof of Theorem~\ref{thm:estimate_bias}}
\label{app:proof:estimate_bias}
\proof{Proof.}
    First, we focus on the denominator of the expression $L^r(x)$:
    \begin{align*}
        \max(F^{NC}_r(t\mid x), F^{C}_r(t\mid x)) &:= \max(\mathbb{P}(T_r < t \mid x), \mathbb{P}(T' < t, D' = r \mid x))\\
        &= \max(\mathbb{P}(T_r < t \mid x), \mathbb{P}(T_r < t, D' = r \mid x)) \tag{By Definition of $T'$}\\
        &= \mathbb{P}(T_r < t \mid x)
    \end{align*}
    
    Then, we can simplify the equality as follows:
    \begin{align*}
        L^r(t, x) &:= \frac{F^{NC}_r(t\mid x) - F^{C}_r(t\mid x)}{\mathbb{P}(T_r < t \mid x)} \\
        &= \frac{\mathbb{P}(T_r < t \mid x) - \mathbb{P}(T' < t, D' = r \mid x)}{\mathbb{P}(T_r < t \mid x)}\\
        &= \frac{\mathbb{P}(T_r < t \mid x) - \mathbb{P}(T_r < t, D' = r\mid x)}{\mathbb{P}(T_r < t \mid x)} \tag{By Definition of $T'$}\\
        &= \mathbb{P}(D' \neq r \mid T_r < t, x) \tag{By Bayes Theorem}
    \end{align*}
\endproof

\section{Simulation study}
\subsection{Data Generation}
\label{app:data_generation}
This section formally describes the data generation used for the simulations presented in Section~\ref{sec:simulation}.

\paragraph{Covariates.} To generate the covariates, we first draw the group membership following a Bernoulli distribution. Then, we generated the two independent covariates following two normal distributions with group-specific centers, and unit variance, specifically $c_1 = (1.5, 1.5)$ and $c_2 = - c_1$. All other covariates were independently drawn from standard normal distributions. Formally, the group and covariates were modeled using the following procedure:
\begin{align*}
    G &\sim \text{Bernoulli}(0.5)\\
    X_{[1, 2]} \mid G = g &\sim \text{MVN}(c_g, I^2)\\
    X_{[3:10]} &\sim \text{MVN}(0, I^8)
\end{align*}
with MVN denoting a multivariate normal distribution, and $I^n$, the identity covariance matrix of dimension $n$.

\paragraph{Competing risks.} From the generated covariates, we then draw two competing risks $R = 2$ following the procedure introduced in~\cite{beyersmann2009simulating}:
\begin{itemize}
    \item Choose the distribution of the cause-specific hazards $\lambda_r$ associated with each competing risk $r$. This quantity corresponds to the instantaneous risk for the event $r$, ignoring the competing risks.
    \item Simulate the time of the first observed event $T'$ following the all-cause hazard equal to the sum of the cause-specific hazards: $\sum_r \lambda_r$.
    \item For each simulated time $T' = t$, draw the associated event type $D' = d$ from a Bernoulli with probability the relative hazard for event $r$: $\frac{\lambda_r(t)}{\sum_s \lambda_s(t)}$.
\end{itemize}

To ensure each individual and each group have different survival profiles, we draw group-specific coefficients $K$ and $\Phi$ from normal distributions, used to parametrize transformations of the covariates defining Gompertz's scale and shape. Formally, each risk $r$ has for cause-specific hazard $\lambda_r$, the form:
$$\lambda_r(t \mid x, g) = w_r(\kappa_r^g, x) \cdot \exp(w_s(\phi^g, x) \times t)$$
with $\kappa_r^g$ and $\phi^g$, realizations of the $K$ and $\Phi$ for group $g$, and $w_r$ and $w_s$ transformations of the covariates $x$ used to ensure the positivity and non-linearity of the Gompertz's scale and shape defined, similarly to~\cite{lee2018deephit}, as:
\begin{align*}
    w_1(\kappa_1^g, x) &= \big|(\kappa_1^g[5:10] \cdot x[5:10])^2 + \kappa_1^g[1:5] \cdot x[1:5]\big| \tag{Shape Cause 1}\\
    w_2(\kappa_2^g, x) &= \big|(\kappa_2^g[1:5] \cdot x[1:5])^2 + \kappa_2^g[5:10] \cdot x[5:10]\big| \tag{Shape Cause 2}\\
    w_s(\phi^g, x) &= \big|\phi^g \cdot x[5:10]\big| \tag{Shift}
\end{align*}
with the notation $v[a:b]$ corresponding at the selection of the dimension $a$ to $b$ of the vector $v$, and $|\cdot|$, the absolute value.

Following the previously described procedure from~\cite{beyersmann2009simulating}, the overall hazard associated with the first observed event ($T'$) is the sum of the cause-specific hazards with associated event type ($D'$) drawn from a trial following the relative risk of each event. 
\begin{align*}
    K^g_1 &\sim \text{MVN}(0, \sigma_K^2 I^{10}) \tag{Group-specific shape for $r = 1$}\\
    K^g_2 &\sim \text{MVN}(0, \sigma_K^2 I^{10}) \tag{Group-specific shape for $r = 2$}\\
    \Phi^g &\sim \text{MVN}(0, \sigma_{\Phi}^2 I^{5}) \tag{Shared scale}\\[1.5em]
    T' &\mid X, G, K^g_1, K^g_2, \Phi^g = (x, g, \kappa_1^g, \kappa_2^g, \phi^g)\\ &\sim \text{Gompertz}\left(w_1(\kappa_1^g, x) + w_2(\kappa_2^g, x), w_s(\phi^g, x)\right)\\
    D' = 1 &\mid X, G, K^g_1, K^g_2 = (x, g, \kappa_1^g, \kappa_2^g) \\&\sim \text{Bernoulli}\left(\frac{w_1(\kappa_1^g, x)}{w_1(\kappa_1^g, x) + w_2(\kappa_2^g, x)}\right)
\end{align*}

\paragraph{Censoring.} To mimic the real-world settings, we generate censoring independent from the two competing risks. We draw censoring times following a Gompertz hazard distribution\footnote{This differs from~\cite{lee2018deephit, nagpal2022counterfactual} to ensure non-informative censoring.} with shape: $w_c(\zeta, x) = (\zeta \cdot x[5:10])^2$ with $\zeta$, the realization of $Z$ a multivariate normal random variable \footnote{In our simulations, we choose $\sigma_K  = \sigma_Z = \sigma_{\Phi} = 1$ for all experiments.} with variance $\sigma_Z$. Observed event times and types ($T, D$) are the minimum between the first observed competing risks time $T'$ and censoring $C$.
\begin{align*}
    Z &\sim \text{MVN}(0, \sigma_Z^2 I^{6})\\
    C \mid X = x, Z = \zeta &\sim \text{Gompertz}(w_c(\zeta, x))\\
    T &= \text{min}(C, T')\\
    D &= \mathbbm{1}(C > T')
\end{align*}
with $\mathbbm{1}$, the indicator function.

\subsection{Derivation of error}
\label{apd:derivation}

In Section~\ref{sec:simulation}, we compare the theoretical estimate to the empirical ones, the following details the computation of the theoretical quantity of interest.

\begin{align*}
    \mathbbm{P}(T_2 < T_1 \mid T_1 < t) &= \frac{\int_0^t \mathbbm{P}(T_2 < s) f_1(s) ds}{\mathbbm{P}(T_1 < t)}\\
    &= \frac{\int_0^t F_2(s) f_1(s) ds}{F_2(t)}
\end{align*}
with $F_i$ the cumulative density function and $f_i$, the associated probability density function for Gompertz distribution. As the previous equation does not present a closed form, we estimate this quantity by numerical integration and average across all generated patients.

Similarly, the group gap can be estimated by averaging the previous quantity in each generated cluster.

\section{Case-study}

\subsection{Discrimination differences}
\label{apd:discrimnation_diff}
In Section~\ref{subsec:disparity}, our analysis focuses on calibration differences. In this section, we compare the differences in td-CI, quantifying the discrimination differences between the different groups of interest. While all groups benefit from modeling competing risks, older patients most benefit from this practice. Of interest, the fairness gap reduces by modeling competing risks, except in shorter time horizons as shown by the sex split. Note that this observation aligns with our simulations that show that the benefit associated with modeling competing risks may be more limited at shorter time horizons.

\begin{table}[!ht]
        \small
        \centering
        \caption{Discrimination differences - Means and standard deviations over 5-fold cross-validation. {Larger values correspond to improved calibration for the competing risk model.} $|\Delta_{\geq}^C| - |\Delta_{\geq}^{NC}|$ corresponds to the reduction of the fairness gap between younger and older than 50 patients when modeling competing risks.}
        \begin{tabular}{cccc}
            \toprule
            \multirow{2}{*}{\textbf{Age}}  & \multicolumn{3}{c}{\textbf{td-CI Difference}}\\
               &            1 year&           5 years&           10 years \\\cmidrule(lr){2-4}
                $<$ 50 & 0.005 (0.008) & 0.002 (0.010) & 0.000 (0.006) \\
                $\geq$ 50  & 0.012 (0.009) & 0.008 (0.007) & 0.005 (0.002) \\
                \cmidrule(lr){2-4}
                 $|\Delta_{\geq}^C| - |\Delta_{\geq}^{NC}|$ & -0.004 (0.009) & -0.004 (0.012) & -0.003 (0.006) \\
                \bottomrule
        \end{tabular}
\end{table}

\begin{table}[!ht]
        \small
        \centering
        \caption{Discrimination differences between sex - Means and standard deviations over 5-fold cross-validation. {Larger values correspond to improved calibration for the competing risk model. $|\Delta_{M}^C| - |\Delta_{M}^{NC}|$ corresponds to the reduction of the fairness gap between men and women from modeling competing risks.}}
        \begin{tabular}{c ccc}
            \toprule
            \multirow{2}{*}{\textbf{Sex}}  & \multicolumn{3}{c}{\textbf{td-CI Difference}}\\
               &            1 year&           5 years&           10 years \\\cmidrule(lr){2-4}
                Male & 0.006 (0.009) & 0.003 (0.007) & 0.002 (0.006) \\
                Female & 0.010 (0.009) & 0.006 (0.005) & 0.003 (0.003)   \\\cmidrule(lr){2-4}
                 $|\Delta_{M}^C| - |\Delta_{M}^{NC}|$ & 0.004 (0.013) & 0.000 (0.008) & -0.003 (0.008) \\
            \bottomrule
        \end{tabular}
\end{table}

\end{APPENDICES}

\end{document}